%% file: main_2nd_review_revise.tex
\documentclass[review,12pt]{elsarticle}

\usepackage{amssymb}

\usepackage{lineno}
\usepackage{amsmath,amsfonts}
\usepackage{algorithmic}
\usepackage{algorithm}
\usepackage{array}
\usepackage[caption=false,font=normalsize,labelfont=sf,textfont=sf]{subfig}
\usepackage{textcomp}
\usepackage{stfloats}
\usepackage{url}
\usepackage{verbatim}
\usepackage{graphicx}

\usepackage{epsfig}
\usepackage{graphicx}
\usepackage{amsmath}
\usepackage{amssymb}
\usepackage{booktabs}
\usepackage{stfloats}
\usepackage{float}
\usepackage{pifont}
\usepackage{enumitem}
\usepackage{multirow}
\usepackage{bbding}
\usepackage{etoolbox}
\usepackage{wrapfig}
\usepackage{graphicx}
\usepackage{subcaption}
\usepackage{ulem}
\usepackage{subcaption}
\usepackage{color, colortbl}
\definecolor{Gray}{gray}{0.9}

\newcommand{\nameshort}[1]{CM-Adapter}
\newcommand{\namelong}[1]{Cross-Modal Adapter}
\newcommand{\cmark}{\ding{51}}

\definecolor{mygreen}{RGB}{114, 210, 126}

\usepackage[capitalize]{cleveref}
\crefname{section}{Sec.}{Secs.}
\Crefname{section}{Section}{Sections}
\Crefname{table}{Table}{Tables}
\crefname{table}{Tab.}{Tabs.}

\journal{Pattern Recognition}

\begin{document}

\begin{frontmatter}



\title{Cross-Modal Adapter for Vision-Language Retrieval}


\author[thu]{Haojun Jiang}
\author[bit]{Jianke Zhang}
\author[thu]{Rui Huang}
\author[thu]{Chunjiang Ge}
\author[thu]{Zanlin Ni}
\author[thu]{Shiji Song}
\author[thu]{Gao Huang \corref{cor1}}

\affiliation[thu]{organization={Department of Automation, BNRist, Tsinghua University},
            postcode={100084}, 
            state={Beijing},
            country={China}}

\affiliation[bit]{organization={School of Computer Science and Technology, Beijing Institute of Technology},
            postcode={100081}, 
            state={Beijing},
            country={China}}

\cortext[cor1]{Corresponding Author}
\ead{jhj20@mails.tsinghua.edu.cn (Haojun Jiang); gaohuang@tsinghua.edu.cn}

\begin{abstract}
Vision-language retrieval is an important multi-modal learning topic, where the goal is to retrieve the most relevant visual candidate for a given text query. 
Recently, pre-trained models, \textit{e.g.}, CLIP, show great potential on retrieval tasks.
However, as pre-trained models are scaling up, fully fine-tuning them on donwstream retrieval datasets has a high risk of overfitting. 
Moreover, in practice, it would be costly to train and store a large model for each task. 
To overcome the above issues, we present a novel \textbf{\namelong{}} for parameter-efficient transfer learning.
Inspired by adapter-based methods, we adjust the pre-trained model with a few parameterization layers.
However, there are two notable differences. 
First, our method is designed for the multi-modal domain.
Secondly, it allows encoder-level implicit cross-modal interactions between vision and language encoders.
Although surprisingly simple, our approach has three notable benefits: 
(1) reduces the vast majority of fine-tuned parameters,
(2) saves training time, and
(3) allows all the pre-trained parameters to be fixed, enabling the pre-trained model to be shared across datasets. 
Extensive experiments demonstrate that, without bells and whistles, our approach outperforms adapter-based methods on image-text retrieval datasets (MSCOCO, Flickr30K) and video-text retrieval datasets (MSR-VTT, DiDeMo, and ActivityNet). 
\end{abstract}



\end{frontmatter}



\input{2nd_review_revise/1_introduction}
\input{2nd_review_revise/2_related}
\input{2nd_review_revise/3_method}
\input{2nd_review_revise/4_exp}

\input{2nd_review_revise/5_conclusion}


\section*{Acknowledgements}
This work is supported in part by National Key R\&D Program of China (2021ZD0140407), the National Natural Science Foundation of China under Grant 62022048 and Guoqiang Institute of Tsinghua.


\bibliographystyle{elsarticle-num} 
\bibliography{reference}






\end{document}

%% file: 2nd_review_revise/1_introduction.tex
\section{Introduction}
Visual content carrier, \textit{e.g.,} images and videos, are playing an essential role in retaining and disseminating information.
Take the video as an example, nowadays, on video platforms like YouTube, users search for the videos they are interested in through natural language queries.
Consequently, accurately returning the most relevant videos based on textual queries has evolved into a pressing research topic, gaining traction in both the industry and academia~\cite{sun2021vsrnet, luo2021clip4clip, yang2024continual, song2024deep, yang2024dgl}.

\begin{figure}[t]
\centering
\includegraphics[scale=0.5]{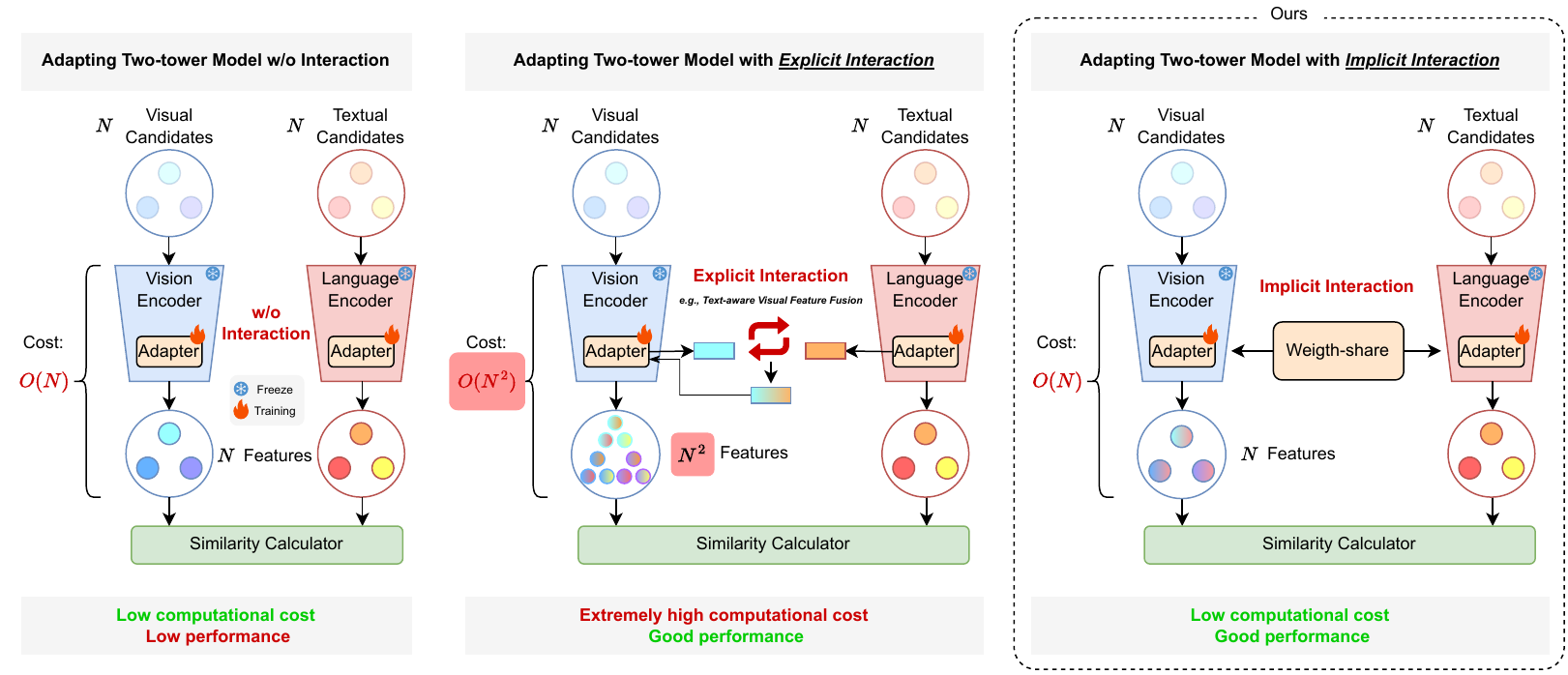}
\caption{
\textbf{Motivation and idea}.
\textit{Left:} The diagram shows a pre-trained two-tower model adapted with vanilla adapters for downstream tasks, where the lack of feature interaction between the encoders results in poor performance.
\textit{Middle:} Introducing explicit feature interaction improves cross-modal alignment but dramatically increases the computational cost to an impractical $O(N^2)$.
\textit{Right:} Our method promotes cross-modal alignment while ensuring reasonable computational cost through a weight-sharing mechanism.
}
\label{fig:1}
\centering
\end{figure}

In the field of vision-language retrieval, an important category of the method is pre-training~\cite{ xu2021videoclip, radford2021learning} that aims at learning better transferable vision-language representations.
A representative work is the Contrastive Language-Image Pre-training model~\cite{radford2021learning}, \textit{i.e.}, CLIP, which exhibits great potential in handling this task.
Numerous works~\cite{luo2021clip4clip, gorti2022x, fang2021clip2video, ma2022x, bain2022clip, gao2021clip2tv, liu2022ts2} have focused on adapting the CLIP to the vision-language retrieval domain.
While these methods deliver promising results across various benchmarks, they require full fine-tuning of the CLIP model and even introduce many new parameters to support cross-modal interactions.
As the current trend of scaling up pre-trained models, this paradigm leads to two issues:
(1) it has a high risk of overfitting on downstream datasets, and 
(2) it is costly to train and store an entirely new large model for each dataset.

An elegant solution to the above-mentioned problems is Adapter~\cite{houlsby2019parameter}, which has achieved great success in natural language processing. 
For example, it attains comparable performance with fully fine-tuned methods by training only 3.6\% parameters.
In the vision-language retrieval task, there are \textit{two modalities} and each with a feature encoder. 
Thus, a straightforward way is to design adapters \textit{separately} for each encoder as shown in \cref{fig:1} (Left).
However, this naive scheme forbids any \textit{encoder-level} cross-modal interaction, and leads to a sub-optimal solution.
Meanwhile, introducing the explicit encoder-level cross-modal interaction inside pre-trained models' encoders through adapters for retrieval tasks is non-trivial. 
Since it requires interactions between every pair of visual candidate and text query, this will change the computational complexity from $O(N)$ to $O(N^2)$ where $N$ is the number of vision-language candidate pairs (\cref{fig:1} Middle).
To be specific, when we aim to construct a vision-language retrieval system, the features of the visual candidates are typically computed offline without the presence of text and reused when different users provide retrieval queries. 
However, if it is now required for the visual features to interact with the text features, this implies that the features of the visual candidates cannot be extracted offline and must be extracted when the user provides the text.
When conducting a retrieval on a vast database, this will substantially increase the computational load, which is untenable in practice.

To address the above challenge, we propose a novel method named \textbf{\namelong{}}. 
The key idea is to enable \textit{encoder-level} cross-modal interactions by sharing adapters' weights between two modalities rather than introducing explicit feature interactions (\cref{fig:1} Right). 
Such a scheme allows an implicit cross-modal interaction, which will facilitate the re-alignment of vision and language feature spaces for the cross-modal retrieval task. 

Furthermore, by adopting this weight-sharing mechanism, computational complexity can be kept as $O(N)$, since each vision or language feature can be obtained independently. 
We conduct extensive experiments on five retrieval datasets, \textit{i.e.}, MSR-VTT~\cite{xu2016msr}, DiDeMo~\cite{anne2017localizing}, ActivityNet~\cite{krishna2017dense}, Flickr30K~\cite{plummer2015flickr30k}, and MSCOCO~\cite{lin2014microsoft}, to demonstrate the effectiveness of our method. 
First, without bells and whistles, \namelong{} attains comparable performance as the fully fine-tuned method~\cite{luo2021clip4clip} while optimizing much fewer parameters, \textit{e.g.}, 0.5M \textit{v.s.} 123.5M.
Secondly, our method saves approximately 30\% of training costs on RTX 3090.
Last, our method outperforms other adapter-based~\cite{he2021towards, pan2022st, chen2022adaptformer} and prompt-based~\cite{zhou2022learning, jia2022visual, khattak2022maple} parameter-efficient methods. 
In summary, this paper makes three-fold contributions:
\begin{itemize}
    \item We propose a parameter-efficient \namelong{} for vision-language retrieval tasks. To the best of our knowledge, we are among the first researchers to investigate adapter-based parameter-efficient transfer learning for the vision-language retrieval domain.
    \item We propose a weight-sharing mechanism to enable encoder-level cross-modal interaction without introducing huge additional computations. 
    \item Experiments show that our approach outperforms adapter-based methods on five vision-language retrieval datasets.
\end{itemize}

%% file: 2nd_review_revise/2_related.tex
\section{Related Work}
\subsection{Vision-Language Retrieval}
Text-video retrieval~\cite{luo2021clip4clip, zhen2019deep, hu2023deep} is becoming one of the most important multi-modal learning topics~\cite{antol2015vqa, yu2018mattnet, johnson2017clevr, zhen2019deep}
with advances in both computer vision~\cite{huang2017densely, he2016deep, dosovitskiy2020image} 
and natural language processing~\cite{vaswani2017attention, brown2020language}.
Recently, great progress has been achieved in the text-video retrieval domain with the advances in pre-training~\cite{sun2021vsrnet, xu2021videoclip, radford2021learning, li2024robust, chiang2022multi}. 
Among these works, Contrastive Language-Image Pre-training model~\cite{radford2021learning}, \textit{i.e.}, CLIP, exhibits great potential on multiple downstream tasks. 
Recently, CLIP4Clip~\cite{luo2021clip4clip} demonstrates that fully fine-tuning the CLIP with a similarity calculator can achieve promising performance on various text-video retrieval datasets. 
As a result, a series of works focus on excavating the power of CLIP by designing parameter-rich cross-modal fusion modules~\cite{gorti2022x,fang2021clip2video,ma2022x,gao2021clip2tv} on top of it or plugging a token selection module~\cite{liu2022ts2} into it.
All these works adopt the fully fine-tuning paradigm following CLIP4Clip. 
However, as the current trend of scaling up the pre-trained models, \textit{e.g.}, the latest BEIT-3~\cite{wang2022image} has 1.9B parameter, such a scheme is easy to overfit, computation-intensive, and time-consuming when applied to downstream datasets. 
Contrary to these works, we make the first attempt to explore parameter-efficient transfer learning on the text-video retrieval domain and achieve comparable performance to fully fine-tuning with much fewer parameters.

Text-image retrieval is an important multimodal learning topic.
The early paradigm is to train a neural network from scratch.
The turning point came in 2019, inspired by the great success in NLP, and there was an explosion of interest in developing large-scale cross-modal pre-training models.
These pre-trained models can be classified into three types: two-stream models, such as CLIP~\cite{radford2021learning} and ALIGN~\cite{yuan2021florence}; 
single-stream models, like OSCAR~\cite{li2020oscar}; 
and hybrid-stream models, such as ALBEF~\cite{li2021align} and BLIP~\cite{li2022blip}. 
Hybrid-stream models, which combine encoder-based and decoder-based methods, are particularly effective at text-image retrieval and support a range of other cross-modal alignment tasks. 
However, adapting these pre-trained models for text-image retrieval is challenging due to the huge training cost for full fine-tuning.
In this paper, we explore an efficient approach to adapt the pre-trained model to this task by inserting a small number of trainable parameters into the model and fixing all pre-trained weights.

\subsection{Parameter Efficient Tuning}
In recent years, pre-trained models~\cite{radford2021learning, yuan2021florence, li2020oscar} have gained significant popularity and prominence. 
However, the process of fine-tuning these models throughout can be computationally intensive and can consume a significant amount of time. 
As a response to this challenge, Parameter Efficient Tuning (PET) was introduced as an innovative approach to adapt pre-trained models more efficiently for downstream tasks. 
Within this context, two major avenues of research have emerged as predominant: adapters~\cite{chen2022conv, pan2022st, he2021towards, chen2022adaptformer} and prompt learning~\cite{zhou2022learning,jia2022visual, khattak2022maple}.

Adapters~\cite{houlsby2019parameter} are streamlined, lightweight layers that are seamlessly integrated into pre-existing, pre-trained models.
During the fine-tuning process, solely the weights associated with the adapters undergo adjustments, while the rest of the model's parameters remain frozen. 
This approach not only facilitates the learning of information specific to downstream tasks but also ensures that the foundational knowledge acquired during the model's initial pre-training stage is retained and undisturbed.
Recently, in the computer vision community, Chen \textit{et~al.}~\cite{chen2022conv} proposes a Conv-Adapter which consists of a point-wise convolution and depth-wise convolution in a bottleneck structure to enable adapters in convolutional neural networks. 
Pan \textit{et~al.}~\cite{pan2022st} further explores image-to-video transfer learning and proposes a spatial-temporal adapter which is able to learn temporal information in videos.
Different from the methods mentioned above, Chen \textit{et~al.}~\cite{chen2022adaptformer} proposes to deploy adapters in the parallel position of certain modules (e.g. MLP residual blocks in VIT). 
While much of the prior research has predominantly focused on single modality, either vision or language in nature, our innovative approach seeks to delve into the intricacies of the multi-modal landscape. 
Specifically, we aim to rigorously evaluate the potential and efficacy of adapters when applied across diverse modalities, bridging the gap between isolated domains.

Prompt learning~\cite{petroni2019language}, originally introduced within the context of NLP, is a technique that leverages pre-trained knowledge for downstream tasks. 
It achieves this by utilizing either human-crafted prompts or prompts that can be learned automatically, effectively probing the model to extract relevant information.
Concretely, a natural language task instruction~(prompt) is prepended in the input, and the whole pre-trained model is frozen. 
The prompt could be discrete or continuous embedding~\cite{petroni2019language,li2021prefix,lester2021power}. 
Since multi-modal pre-trained models, \textit{e.g.}, CLIP~\cite{radford2021learning}, have shown impressive performance on various tasks, context optimization~\cite{zhou2022learning}~(CoOp), has been proposed to prompt CLIP on image classification tasks. 
CoOp optimizes continuous learnable prompts for the text encoder to learn better label distribution for image classification tasks. 
Visual prompt tuning~\cite{jia2022visual}~(VPT) leverages prompt in each layer of vision models. 
Recently, a multi-modal prompt tuning method, MaPLe~\cite{khattak2022maple}, has been proposed to enable cross-modal communication with explicit feature interactions for image classification tasks.
Explicit feature interaction mechanisms are impractical for cross-modal retrieval tasks due to tremendously growing computational demands.
Besides, VoP~\cite{huang2022vop} is a prompt learning method proposed for cross-modal retrieval which focuses on obtaining better prompt features only inside the vision encoder.
In distinction to the aforementioned research approaches, our methodology centers around the use of adapters. 
Empirical evidence from experiments clearly indicates that our method consistently surpasses prompt-based methods.

%% file: 2nd_review_revise/3_method.tex
\section{Method}
\begin{figure*}[t]
\centering
\includegraphics[scale=0.69]{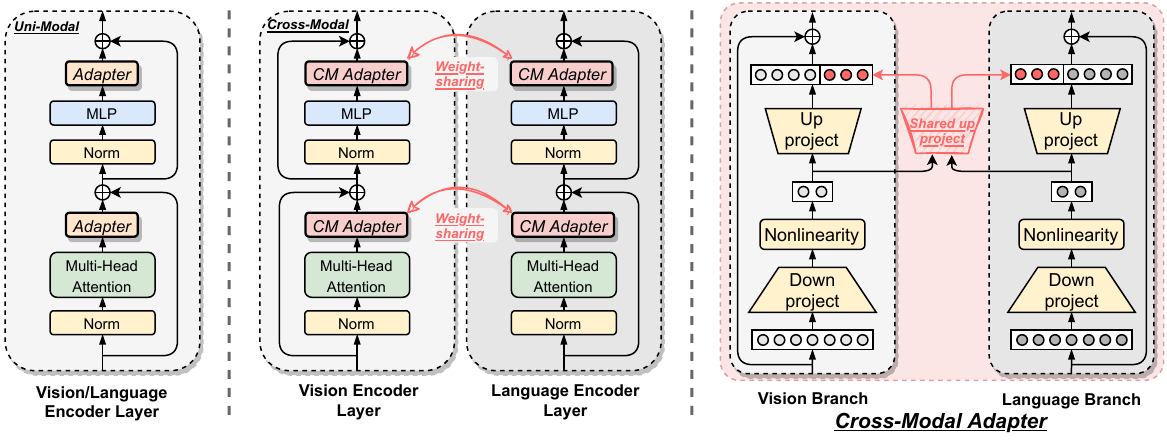}
\caption{\textbf{Overview of the Cross-Modal Adapter.} \textit{Left} is the scheme that naively plugs independent uni-modal adapters into the vision and language encoders. Note that there is no cross-modal interaction. \textit{Middle} is our cross-modal adapter which enables encoder-level cross-modal interaction via a weight-sharing mechanism. \textit{Right} is the implementation details of our cross-modal adapter. }
\label{fig:overview}
\centering
\vspace{-15pt}
\end{figure*}

In this section, we begin with a review of adapting the pre-trained CLIP~\cite{radford2021learning} to retrieval tasks in \cref{CLIP}. 
Then, we introduce our proposed cross-modal adapter that facilities cross-modal interaction with negligible trainable parameters in \cref{CMAdapter}. 
Finally, we illustrate the vision-language matching process in \cref{Simcal}. 
The overall framework is illustrated in \cref{fig:overview}.

\subsection{Preliminary}
\label{CLIP}
\textbf{Problem Formulation.}
Given a set of videos $\mathcal{V}=\left\{v_1,v_2,\ldots,v_n\right\}$ and its corresponding text queries $\mathcal{T}=\left\{t_1,t_2,\ldots,t_n\right\}$, we perform text-video retrieval by calculating the similarity between the text query $t_j$ and each video $v_i$ which consists of $\left|v_i\right|$ frames $\left\{v_{i,1}, v_{i,2}, \ldots, v_{i,|v_i|}\right\}$, and return the video with the highest similarity score. 
The task of text-image retrieval can be considered a special case of text-video retrieval, specifically when each video $v_i$ consists of only one frame $\left\{v_{i,1}\right\}$.
Thus, in the following sections, we primarily use video as an example for explanation.

\textbf{Revisiting CLIP.}
Following recent works~\cite{luo2021clip4clip, huang2022vop}, we adopt the pre-trained CLIP model as the frozen backbone.
Since the CLIP model was trained on a large-scale dataset of text-image pairs, it exhibits strong cross-modal alignment capabilities.
Specifically, consider the video $v_i$, the $j$-th frame $\mathbf{v}_{i,j} \in \mathbb{R}^{H \times W \times C}$ is first split into non-overlapping patches with a size of $P \times P$.
Then the patches are mapped to embeddings with a linear projection and added with learnable positional embeddings.
Besides, a learnable [CLS] token, which represents the global information of the frame, is added at the start of the sequence. 
Formally, we denote the input and features of the vision encoder as $\mathbf{p}_{i,j,l} \in \mathbb{R}^{(M + 1) \times D_v}$, where subscript $l$ means the layer index, $M$ is the total number of patches, and $D_v$ is the hidden dimension. 
The input of the vision encoder is denoted as $\mathbf{p}_{i,j,0}$.

Inside the vision encoder, each layer consists of a self-attention module (MSA), a feedforward MLP, and layer normalization (LN). 
For $j$-th frame of video $v_i$, the modules in $l$-th layer perform calculations as follows:
\begin{align}
\setlength{\abovedisplayskip}{0pt}
\setlength{\belowdisplayskip}{0pt}
    \mathbf{\hat{p}}_{i,j,l} &= \mathrm{MSA}(\mathrm{LN}(\mathbf{p}_{i,j,l-1})) + \mathbf{p}_{i,j,l-1},\ l = 1,...,L,\\[5pt]
    \mathbf{p}_{i,j,l} &= \mathrm{MLP}(\mathrm{LN}(\mathbf{\hat{p}}_{i,j,l})) + \mathbf{\hat{p}}_{i,j,l}, \ \ \ \ \ \mathbf{p}_{i,j,l} \in \mathbb{R}^{D_v}.
\end{align}
After the patches have been processed by all the layers, we regard the feature of the [CLS] token as the global representation for a frame. The final frame representation is obtained as follows:
\begin{align}
\setlength{\abovedisplayskip}{0pt}
\setlength{\belowdisplayskip}{0pt}
    \boldsymbol{f}_{i,j} &= \mathrm{Proj}(\mathrm{LN}(\mathbf{p}_{i,j,L}^{0})),\ \ \ \ \ \ \ \ \ \boldsymbol{f}_{i,j} \in \mathbb{R}^{D_t},
\end{align}
\noindent where $\mathrm{Proj}(\cdot)$ is a linear projection layer to handle the dimension mismatch between video and text encoders, and $D_t$ is the hidden dimension of the text encoder.
Therefore, for video $v_i$, we can get all the frame representations as $\left\{\boldsymbol{f}_{i,1}, \boldsymbol{f}_{i,2}, \ldots, \boldsymbol{f}_{i,|v_i|}\right\}$. 
The video feature is generated based on the given text query feature, as detailed in \cref{Simcal}. 

As for text queries, we adopt CLIP's language encoder to generate their features.
Like the vision encoder, the input text query $t_i \in \mathcal{T}$ is first transformed into a sequence of embeddings by the language encoder. 
Then, a [CLS] token and a [SEP] token are added to the start and end of the sequence, respectively. 
Subsequently, this sequence is going through all the layers of the text encoder. 
Finally, we adopt the feature of [SEP] token at the last layer as global representation $\boldsymbol{t}_i \in \mathbb{R}^{D_t}$ of a text query. 

\subsection{Cross-Modal Adapter}
\label{CMAdapter}
CLIP~\cite{radford2021learning} has shown great potential for transferring to downstream tasks. 
Thus, previous works~\cite{luo2021clip4clip, gorti2022x, liu2022ts2} mainly focus on adapting the powerful pre-trained CLIP to the vision-language retrieval domain by fully fine-tuning the two encoders. 
However, as the pre-trained models are scaling up quickly, this paradigm has a high risk of overfitting and is costly. 
Inspired by recent parameter-efficient transfer learning works~\cite{houlsby2019parameter} in NLP, we proposed a novel Cross-Modal Adapter that enables implicit cross-modal interaction between encoders with negligible trainable parameters for retrieval tasks. 

\textbf{Uni-modal Adapter.} Our method is built based on Adapter~\cite{houlsby2019parameter} which is designed for parameter-efficient transfer learning in NLP. The authors adapt the large pre-trained model to downstream tasks by inserting a few learnable adapter modules inside each transformer layer. It attains good performance by fine-tuning only adapter modules, which contain merely 3.6\% of the pre-trained model's parameters. Specifically, the adapter module follows a bottleneck design to reduce the number of parameters, which consists of a down-projection linear layer, a non-linear layer, and an up-projection layer. 

Formally, given an input feature $\mathbf{x} \in \mathbb{R}^{1 \times d}$, the feed-forward function of the adapter module can be written as:
\begin{equation}
\setlength{\abovedisplayskip}{0pt}
\setlength{\belowdisplayskip}{0pt}
\text{Adapter}(\mathbf{x}) = \mathbf{x} + \sigma(\mathbf{x}\boldsymbol{\mathrm{W}}_{\text{down}})\boldsymbol{\mathrm{W}}_{\text{up}},
\end{equation}
where $\boldsymbol{\mathrm{W}}_{\text{down}} \in \mathbb{R}^{d \times r}$, $\boldsymbol{\mathrm{W}}_{\text{up}} \in \mathbb{R}^{r \times d}(r \ll d)$ are down-projection and up-projection weights respectively, and $\sigma(\cdot)$ is non-linear layer which is implemented as GELUs. Besides, the adapter is initialized as a near-identity mapping to stabilize the training process.

\textbf{\namelong{}.} 
A straightforward way to achieve parameter-efficient transfer learning for the multi-modal model~\cite{radford2021learning} is plugging independent adapters inside each encoder as shown in \cref{fig:overview} (left). 
However, this naive scheme does not facilitate cross-modal communication between vision and language encoders, which might lead to a sub-optimal solution.
Previous work~\cite{miech2021thinking} shows a model with encoder-level cross-modal interaction boosts retrieval performance significantly. 
Furthermore, we empirically demonstrate such a naive scheme is sub-optimal, as shown in \cref{tab:ab_share}. 
Meanwhile, introducing the cross-modal interaction among shallow layers in CLIP is non-trivial. 
It requires interactions between every pair of visual content and text query, which leads to quadratic computational costs with respect to the sample number. 
In practice, such a surge in computational cost will lead to a slow retrieval process and a bad user experience.

To address the above challenge, we propose a novel \namelong{} to enable implicit encoder-level cross-modal interaction method via a weight-sharing mechanism as shown in \cref{fig:overview} (middle). 
Intuitively, shared parameters project inputs into a unified feature space, which promotes cross-modal alignment. 
Furthermore, these shared parameters allow gradients from the vision encoder to be propagated to the language encoder, thereby accelerating the alignment of features.
This design avoids introducing huge amounts of computation during inference since there is no explicit feature interaction.

In \namelong{}, the weight-sharing mechanism is applied in the up-projection layer of adapters, as shown in \cref{fig:overview} (right). 
Empirically, we find that the up-projection layer inside the vanilla adapter is more likely to learn modality-specific features that can impair cross-modal alignment.
Furthermore, in \cref{tab:ab_sharepos}, we demonstrate that sharing weight in the up-projection layer outperforms all other strategies, \textit{i.e.}, in down-projection or both.

Formally, given an input feature $\mathbf{x} \in \mathbb{R}^{1 \times d}$, where $\mathbf{x}$ can be a patch feature from an image or a video frame, or a token feature from a text query, the bottleneck feature $\mathbf{z} \in \mathbb{R}^{1 \times r}$ can be obtained as follows:
\begin{equation}
\setlength{\abovedisplayskip}{0pt}
\setlength{\belowdisplayskip}{0pt}
\mathbf{z} = \sigma(\mathbf{x}\boldsymbol{\mathrm{W}}_{\text{down}}),
\end{equation}
where $\boldsymbol{\mathrm{W}}_{\text{down}} \in \mathbb{R}^{d \times r}$ and $\sigma(\cdot)$ is non-linear layer. Then, the up-projection layer with a weight-sharing mechanism and the output of the cross-modal adapter can be written as:
\begin{equation}
\setlength{\abovedisplayskip}{0pt}
\setlength{\belowdisplayskip}{0pt}
\text{Adapter}_{\text{CM}}(\mathbf{x}) = \mathbf{x} + \text{Concat} \left[\mathbf{z}\boldsymbol{\mathrm{W}}_{\text{up,unique}}, \text{ } \mathbf{z}\boldsymbol{\mathrm{W}}_{\text{up,share}} \right],
\end{equation}
where $\boldsymbol{\mathrm{W}}_{\text{up,share}} \in \mathbb{R}^{r \times d_{s}}, 0 < d_{s} \leq d$ is modality-shared weights while $\boldsymbol{\mathrm{W}}_{\text{up,unique}} \in \mathbb{R}^{r \times (d - d_{s})}$ is modality-specific weights. Finally, cross-modal adapters are inserted after self-attention and feedforward MLP modules.

\subsection{Vision-Language Matching}
\label{Simcal}
After extracting all the features, the last essential step is calculating the similarity between every pair of visual candidate and text query. 
For image-text retrieval, we directly calculate the cosine distance of image and text feature as their similarity.
As for video-text retrieval, in contrast to the previous work~\cite{gorti2022x} that introduces a parameter-rich similarity calculator, we apply a parameter-free one~\cite{bain2022clip} to minimize the number of fine-tuned parameters. 
First, for videos, we generate query-aware video features, which aggregate important information from each frame based on the given text query feature. 
Concretely, we calculate the inner product $\alpha_{j}$ between the text query feature $\boldsymbol{t}$ and $j$-th frame feature $\boldsymbol{f}_{i,j}$ of video $v_i$. 
Formally, we have the following equation:
\begin{equation}
\setlength{\abovedisplayskip}{0pt}
\setlength{\belowdisplayskip}{0pt}
\alpha_{j} = \langle \, \boldsymbol{t}, \; \boldsymbol{f}_{i,j} \rangle, \ \ \ j \in \{1,\ldots, \left|v_i\right|\},
\end{equation}
where $\langle\cdot, \cdot\rangle$ represents inner product. 
Then, we obtain a final query-aware embedding $\boldsymbol{\hat{v}}_i$ for the video $v_i$ by weighted averaging all video frame features:
\begin{equation}
\setlength{\abovedisplayskip}{0pt}
\setlength{\belowdisplayskip}{0pt}
\hat{\alpha}_j=\frac{e^{\alpha_j / \tau}}{\sum_{k \in \{1,\ldots, \left|v_i\right|\}} e^{\alpha_k / \tau}}, \ \ \boldsymbol{\hat{v}}_{i}=\sum_{j \in \{1,\ldots, \left|v_i\right|\}} \hat{\alpha}_j \boldsymbol{f}_{i,j},
\end{equation}
where $\tau$ is the temperature hyper-parameter, which determines the way of average weighting. 
For instance, if $\tau$ is very small, the embedding of the whole video will be almost the same as that of the most relevant frame to the query feature $\boldsymbol{t}$. 
Contrarily, if $\tau$ is very large, this simply becomes a uniform average without weights. 
Though its simplicity, this parameter-free aggregation method can effectively perform the \textit{temporal modeling} for the video by considering different degrees of text-frame relevance. 

For a batch with $n$ text-video  (or text-image) pairs, we calculate the similarities between all the visual features and all the text query features, resulting in a $n \times n$ similarity score map. 
Then, we adopt a symmetric cross-entropy loss. Formally, we have the following equations:
\begin{align}
    \setlength{\abovedisplayskip}{0pt}
    \setlength{\belowdisplayskip}{0pt}
    \mathcal{L}_{t 2 v} &=-\frac{1}{n} \sum_i^n \log \frac{\exp \left(s\left(\boldsymbol{\hat{v}}_i, \boldsymbol{t}_i\right)\right)}{\sum_{j=1}^n \exp \left(s\left(\boldsymbol{\hat{v}}_j, \boldsymbol{t}_i\right))\right.}, \\[3pt]
    \mathcal{L}_{v 2 t} &=-\frac{1}{n} \sum_i^n \log \frac{\exp \left(s\left(\boldsymbol{\hat{v}}_i, \boldsymbol{t}_i\right)\right)}{\sum_{j=1}^n \exp \left(s\left(\boldsymbol{\hat{v}}_i, \boldsymbol{t}_j\right))\right.},
\end{align}
\begin{equation}
\mathcal{L} =\frac{1}{2}(\mathcal{L}_{v 2 t}+\mathcal{L}_{t 2 v}),
\end{equation}
where $s(\cdot, \cdot)$ denotes cosine distance.

%% file: 2nd_review_revise/4_exp.tex
\section{Experiments}
We first introduce the tasks and datasets adopted for evaluating our method, followed by implementation details, and evaluation metrics.
\textbf{Datasets.} We evaluate our method on two text-image retrieval datasets: Flickr30K~\cite{plummer2015flickr30k} and MSCOCO~\cite{lin2014microsoft}.
For text-video retrieval, three datasets are used: MSR-VTT~\cite{xu2016msr}, DiDeMo~\cite{anne2017localizing}, and ActivityNet~\cite{krishna2017dense}.

\textbf{Evaluation metrics.} For the text-video retrieval task, we use metrics: recall at rank K (R@K, higher is better), MnR (mean rank, lower is better), and MdR (median rank, lower is better) following \cite{luo2021clip4clip}.
For the text-image retrieval task, we report R@1, R@5, and R@10 following \cite{li2021align}.

\textbf{Implementation details.} 
Following CLIP4Clip~\cite{luo2021clip4clip}, we adopt the vision encoder and language encoder from the pre-trained CLIP (ViT-B/32). 
The cross-modal adapter is applied in every layer of two encoders.
We set the bottleneck dimension $r$ as 8 and weight-sharing dimension $d_s$ as 16 by default. 
For ActivityNet, we set the bottleneck dimension $r$ as 16.
Inside the cross-modal adapter, we apply a GELU approximation as the non-linear layer  because it calculates faster. 
Employing the default weight-sharing dimension of 16 surpasses the baselines across all datasets and performs even better with dataset-specific weight-sharing dimension. 
We report the best performance in this paper.
For text-video retrieval datasets~\cite{anne2017localizing, xu2016msr, krishna2017dense}, we first sample frames with 1 FPS for videos. 
Then, uniformly sample 12 frames for MSR-VTT, while 64 frames for ActivityNet and DiDeMo. 
Before training, we initialize the Cross-modal Adapter with a normal distribution . The initial learning rate is 1e-5 for text-video datasets and 1e-3 for text-image datasets. 
Warm-up is applied in the first 10\% of the training process following \cite{luo2021clip4clip}, after which the learning rate is gradually reduced using a cosine decay schedule. 
The batch size is set to 128 for text-video datasets and 256 for text-image datasets due to the limitation of GPU memory.

\subsection{Comparison to the State-of-the-Arts}
Compared with full-tuning baseline~\cite{luo2021clip4clip}, our method reduces 99.6\% trained parameters with even better performance on DiDeMo (\cref{tab:main_didemo}), MSR-VTT (\cref{tab:main_msrvtt}), and Flickr30K (\cref{tab:main_image}). 
The reason for this phenomenon is that these datasets have very limited training samples, \textit{e.g.,} DiDeMo only has 8 thousands samples, while MSCOCO has over half million image-text pairs.
Obviously, fine-tuning with large parameters on small datasets has a high risk of overfitting.
Conversely, our approach requires optimization of just 0.5M parameters, making it particularly well-suited for scenarios involving smaller datasets.

Compared with prompt-based methods~\cite{huang2022vop, zhou2022learning, jia2022visual, khattak2022maple}, our method largely improves performance with comparable parameters.
First, the uni-modal prompt learning methods, \textit{i.e.,} CoOp and VPT, exhibit quite poor performances, which indicates that relying solely on uni-modal prompts is less than ideal when tackling multi-modal tasks.
Secondly, the multi-modal prompt-based approach, VL-Prompt, shows enhanced performance when compared to the uni-modal prompt techniques. 
However, it still lags behind the results achieved by our method.
Besides, the absence of cross-modal interaction in VoP also leads to its poor performance.
Lastly, the most recent advancement in multi-modal prompt learning, namely MaPLe, has incorporated a cross-modal interaction mechanism between visual and textual prompts which introduces a significantly larger number of parameters. 
Even though it fine-tunes a large number of parameters (4.8M vs. 0.5M), it still worse than ours. 
Although prompt learning achieves great success in uni-modal tasks, it is incapable of tackling the challenging multi-modal task. 

\begin{table}[!tp]\footnotesize
\caption{Text-video retrieval results on DiDeMo. 
}
\vspace{-5pt}
\centering
\resizebox{\columnwidth}{!}{
\begin{tabular}{p{3cm} r | p{0.9cm}<{\centering} p{0.9cm}<{\centering} p{0.9cm}<{\centering} p{0.9cm}<{\centering} p{0.9cm}<{\centering} | p{0.9cm}<{\centering} p{0.9cm}<{\centering} p{0.9cm}<{\centering} p{0.9cm}<{\centering} p{0.9cm}<{\centering}}
\toprule
\multirow{2}{*}{\begin{tabular}[c]{@{}c@{}} Methods \end{tabular}} & Trained & \multicolumn{5}{c|}{Text $\Longrightarrow$ Video} & \multicolumn{5}{c}{Vide $\Longrightarrow$ Text}  \\ 
 & Params & R@1↑ & R@5↑ & R@10↑ & MnR↓ & MdR↓ & R@1↑ & R@5↑ & R@10↑ & MnR↓ & MdR↓ \\ 
\midrule
Full-tuning~\cite{luo2021clip4clip} & 123.5M & 43.4 & 70.2 & 80.6 & 17.5 & 2.0 & 42.5 & 70.6 & 80.2 & 16.1 & 2.0 \\
\midrule
Partial~\cite{jia2022visual} & 10.2M & 37.0 & 64.1 & 75.1 & 25.2 & 3.0 & 36.7 & 63.9 & 75.5 & 16.1 & 3.0 \\
Projection~\cite{jia2022visual} & 0.7M & 35.6 & 61.3 & 72.6 & 24.4 & 3.0 & 34.5 & 60.9 & 72.6 & 18.8 & 3.0  \\
\textbf{Prompt-based:} & & & & & & & \\
CoOp~\cite{zhou2022learning} & 0.002M & 32.5 & 60.4 & 71.7 & 29.2 & 3.0 & 34.9 & 63.3 & 75.0 & 16.1 & 3.0 \\
VPT~\cite{jia2022visual} & 0.05M & 35.0 & 61.2 & 71.7 & 29.9 & 3.0 & 33.3 & 60.1 & 70.7 & 19.8 & 3.0 \\
VL-Prompt & 0.08M & 35.3 & 64.1 & 72.8 & 24.0 & 3.0 & 34.2 & 62.4 & 73.7 & 17.1 & 3.0 \\
MaPLe~\cite{khattak2022maple} & 4.8M & 37.5 & \underline{65.9} & 75.8 & 20.8 & 3.0 & 36.4 & \underline{65.6} & 74.8 & \underline{13.8} & 3.0 \\
VoP~\cite{huang2022vop} & 14.3M & \underline{40.0} & 68.0 & \underline{78.5} & \underline{18.3} & 2.0 & \underline{39.1} & 65.3 & \underline{76.7} & \underline{13.8} & 3.0 \\
\textbf{Adapter-based:} & & & & & & & &\\
$\text{Adapter-ATTN}$~\cite{he2021towards} & 2.0M & 36.4 & 62.8 & 73.9 & 23.5 & 3.0 & 36.3 & 64.4 & 74.8 & 15.4 & 2.0 \\ 
$\text{AdaptFormer}$~\cite{chen2022adaptformer} & 2.0M & 36.3 & 63.4 & 75.4 & 22.9 & 3.0 & 35.6 & 64.3 & 75.6 & 14.8 & 3.0 \\ 
$\text{ST-Adapter}$~\cite{pan2022st} & 7.2M & 38.4 & 65.3 & 76.2 & 22.5 & 2.0 & 35.6 & 63.5 & 73.9 & 16.0 & 2.0 \\
\midrule
\textbf{Ours} & \textbf{0.5M} & \textbf{45.6} & \textbf{71.8} & \textbf{80.8} & \textbf{16.2} & \textbf{2.0} & \textbf{44.6} & \textbf{71.4} & \textbf{82.2} & \textbf{9.6} & \textbf{2.0} \\
\bottomrule
\end{tabular}
}
\label{tab:main_didemo}
\end{table}

\begin{table}[!htp]\footnotesize
\caption{Text-video retrieval results on MSR-VTT. 
}
\vspace{-5pt}
\centering
\resizebox{\columnwidth}{!}{
\begin{tabular}{p{3cm} r | p{0.9cm}<{\centering} p{0.9cm}<{\centering} p{0.9cm}<{\centering} p{0.9cm}<{\centering} p{0.9cm}<{\centering} | p{0.9cm}<{\centering} p{0.9cm}<{\centering} p{0.9cm}<{\centering} p{0.9cm}<{\centering} p{0.9cm}<{\centering}}
\toprule
\multirow{2}{*}{\begin{tabular}[c]{@{}c@{}} Methods \end{tabular}} & Trained & \multicolumn{5}{c|}{Text $\Longrightarrow$ Video} & \multicolumn{5}{c}{Vide $\Longrightarrow$ Text}  \\ 
 & Params & R@1↑ & R@5↑ & R@10↑ & MnR↓ & MdR↓ & R@1↑ & R@5↑ & R@10↑ & MnR↓ & MdR↓ \\ 
\midrule
Full-tuning~\cite{luo2021clip4clip} & 123.5M & 43.1 & 70.4 & 80.8 & 16.2 & 2.0 & 43.1 & 70.5 & 81.2 & 12.4 & 2.0 \\
\midrule
Partial~\cite{jia2022visual} & 10.2M & \underline{43.1} & 69.4 & 79.4 & 17.2 & 2.0 & 42.9 & 70.8 & 81.3 & 12.3 & 2.0 \\
Projection~\cite{jia2022visual} & 0.7M & 37.1 & 63.0 & 76.1 & 19.2 & 2.0 & 37.2 & 64.6 & 75.9 & 16.3 & 2.0 \\
\textbf{Prompt-based:} & & & & & & & \\
CoOp~\cite{zhou2022learning} & 0.002M & 38.6 & 63.5 & 74.4 & 18.0 & 3.0 & 41.9 & 68.0 & 78.4 & 11.6 & 2.0 \\
VPT~\cite{jia2022visual} & 0.05M & 40.6 & 66.2 & 75.3 & 19.2 & 2.0 & 40.5 & 66.1 & 76.8 & 17.4 & 2.0 \\
VL-Prompt & 0.08M & 42.0 & \underline{70.7} & \underline{80.2} & 13.7 & 2.0 & 42.5 & \underline{71.9} & \underline{81.4} & 10.1 & 2.0 \\
MaPLe~\cite{khattak2022maple} & 4.8M & 42.6 & 69.6 & 80.1 & \textbf{14.0} & 2.0 & \underline{43.1} & 71.7 & 81.2 & \textbf{9.6} & 2.0 \\
VoP~\cite{huang2022vop} & 14.3M & 40.8 & 68.1 & 79.0 & 15.8 & 2.0 & 42.3 & 70.1 & 81.1 & 11.4 & 2.0 \\
\textbf{Adapter-based:} & & & & & & & &\\
$\text{Adapter-ATTN}$~\cite{he2021towards} & 2.0M & 37.6 & 63.2 & 75.8 & 18.7 & 3.0 & 37.9 & 66.1 & 77.4 & 14.7 & 2.0 \\ 
$\text{AdaptFormer}$~\cite{chen2022adaptformer} & 2.0M & 38.2 & 63.5 & 76.4 & 17.9 & 3.0 & 39.9 & 66.8 & 77.7 & 14.2 & 2.0 \\ 
$\text{ST-Adapter}$~\cite{pan2022st} & 7.2M & 40.0 & 65.5 & 75.8 & 19.2 & 2.0 & 40.1 & 66.4 & 76.6 & 17.4 & 2.0 \\
\midrule
\textbf{Ours} & \textbf{0.5M} & \textbf{45.0} & \textbf{72.3} & \textbf{81.0} & \underline{14.6} & \textbf{2.0} & \textbf{44.6} & \textbf{72.1} & \textbf{83.6} & \underline{9.8} & \textbf{2.0} \\
\bottomrule
\end{tabular}
}
\label{tab:main_msrvtt}
\end{table}

\begin{table}[!t]\footnotesize
\caption{Text-video retrieval results on ActivityNet.}
\vspace{-5pt}
\centering
\resizebox{\columnwidth}{!}{
\begin{tabular}{p{3cm} r | p{0.9cm}<{\centering} p{0.9cm}<{\centering} p{0.9cm}<{\centering} p{0.9cm}<{\centering} p{0.9cm}<{\centering} | p{0.9cm}<{\centering} p{0.9cm}<{\centering} p{0.9cm}<{\centering} p{0.9cm}<{\centering} p{0.9cm}<{\centering}}
\toprule
\multirow{2}{*}{\begin{tabular}[c]{@{}c@{}} Methods \end{tabular}} & Trained & \multicolumn{5}{c|}{Text $\Longrightarrow$ Video} & \multicolumn{5}{c}{Video $\Longrightarrow$ Text}  \\ 
 & Params & R@1↑ & R@5↑ & R@10↑ & MnR↓ & MdR↓ & R@1↑ & R@5↑ & R@10↑ & MnR↓ & MdR↓ \\ 
\midrule
Full-tuning~\cite{luo2021clip4clip} & 123.5M & 41.4 & 71.2 & 82.5 & 8.0 & 2.0 & 41.6 & 72.3 & 84.8 & 7.5 & 2.0 \\
\midrule
Partial~\cite{jia2022visual} & 10.2M & 34.7 & 64.8 & 77.6 & 10.9 & 3.0 & 33.2 & 64.5 & 78.4 & 10.0 & 3.0 \\
Projection~\cite{jia2022visual} & 0.7M & 30.0 & 58.7 & 72.9 & 14.9 & 4.0 & 28.8 & 57.8 & 72.8 & 13.5 & 4.0 \\
\textbf{Prompt-based:} & & & & & & & \\
CoOp~\cite{zhou2022learning} & 0.002M & 27.8 & 56.4 & 69.9 & 18.3 & 4.0 & 29.4 & 58.3 & 72.6 & 13.7 & 4.0 \\
VPT~\cite{jia2022visual} & 0.05M & 30.6 & 59.4 & 72.9 & 16.5 & 4.0 & 28.6 & 57.9 & 71.8 & 14.6 & 4.0 \\
VL-Prompt & 0.08M & 33.2 & 62.9 & 76.1 & 11.6 & 3.0 & 31.1 & 62.6 & 76.7 & 10.8 & 3.0 \\
MaPLe~\cite{khattak2022maple} & 4.8M & 30.8 & 61.0 & 74.2 & 13.3 & 3.0 & 31.0 & 62.2 & 75.7 & 11.4 & 3.0  \\
VoP~\cite{huang2022vop} & 14.3M & 32.6 & 62.5 & 76.5 & 12.0 & 3.0 & 34.2 & 64.8 & 78.4 & 10.7 & 3.0  \\ 
\textbf{Adapter-based:} & & & & & & & \\
$\text{Adapter-ATTN}$~\cite{he2021towards} & 2.0M & 39.0 & \underline{70.4} & 82.1 & 9.0 & 2.0 & \underline{38.2} & \textbf{71.0} & 83.2 & \underline{7.9} & 2.0   \\ 
AdaptFormer~\cite{chen2022adaptformer} & 2.0M & \underline{39.4} & \underline{70.4} & \underline{82.4} & \underline{8.3} & 2.0 & 37.6 & \underline{70.4} & \underline{83.3} & \textbf{7.3} & 2.0  \\ 
ST-Adapter~\cite{pan2022st} & 7.2M & 34.0 & 62.9 & 76.6 & 12.5 & 3.0 & 32.9 & 63.7 & 78.0 & 10.9 & 3.0 \\
\midrule
\textbf{Ours} & \textbf{1.0M} & \textbf{40.9} & \textbf{71.6} & \textbf{82.7} & \textbf{8.0} & \textbf{2.0} & \textbf{39.4} & \textbf{71.0} & \textbf{83.9} & \textbf{7.3} & \textbf{2.0} \\
\bottomrule
\end{tabular}
}
\label{tab:main_actnet}
\end{table}

\begin{table}[!htp]\small
\caption{Text-image retrieval results on Flickr30K and MSCOCO.}
\vspace{-5pt}
\centering
\resizebox{\columnwidth}{!}{
\begin{tabular}{p{3cm} r | p{.7cm}<{\centering} p{.7cm}<{\centering} p{1cm}<{\centering} | p{.7cm}<{\centering} p{.7cm}<{\centering} p{1cm}<{\centering} | p{.7cm}<{\centering} p{.7cm}<{\centering} p{1cm}<{\centering} | p{.7cm}<{\centering} p{.7cm}<{\centering} p{1cm}<{\centering}}
\toprule
& & \multicolumn{6}{c|}{Flickr30K} & \multicolumn{6}{c}{MSCOCO} \\
\midrule
\multirow{2}{*}{\begin{tabular}[c]{@{}c@{}} Methods \end{tabular}} & Trained & \multicolumn{3}{c|}{TR} & \multicolumn{3}{c|}{IR} & \multicolumn{3}{c|}{TR} & \multicolumn{3}{c}{IR} \\ 
 & Params & R@1↑ & R@5↑ & R@10↑ & R@1↑ & R@5↑ & R@10↑ & R@1↑ & R@5↑ & R@10↑ & R@1↑ & R@5↑ & R@10↑ \\  
\midrule
Full-tuning~\cite{luo2021clip4clip} & 123.5M & 88.3 & 98.0 & 99.2 & 75.7 & 94.1 & 96.7 & 66.0 & 88.0 & 93.4 & 49.2 & 76.6 & 85.3 \\
\midrule
Partial~\cite{jia2022visual} & 10.2M & 81.7 & 95.5 & 97.9 & 66.5 & 89.4 & 94.0 & 57.2 & 83.3 & 90.5 & 42.7 & 71.4 & 81.5 \\
Projection~\cite{jia2022visual} & 0.7M &82.7 &96.9 &99.0 & 66.9&88.9 &94.1 &55.1 &81.5 &89.2 &40.0 &68.3 &78.7 \\
\textbf{Prompt-based:} & & & & & & & & & & & \\
CoOp~\cite{zhou2022learning} &0.002M &82.3 &96.1 & 98.1& 64.1&88.3 &93.1  &53.7 &77.9 &86.1 &36.1 &62.2 &72.9  \\
VPT~\cite{jia2022visual} & 0.05M &83.4 &96.4 &98.9 &69.3 &90.5 &94.9 & 55.5& 79.5& 87.9&40.1 &67.0 &77.4 \\
VL-Prompt & 0.08M & 87.3 & 97.7 & 99.2 & 72.7 & 92.3 & 96.1 &59.8 &83.2 &90.5 &43.6 &70.8 &80.4 \\
MaPLe~\cite{khattak2022maple} & 4.8M &\underline{87.5} &97.6 &99.1 &72.6 &91.9 &96.0 &59.8 &82.8 &89.9 &43.4 &70.8 &80.3 \\
\textbf{Adapter-based:} & & & & & & & & & & & \\
AdapterATTN~\cite{he2021towards} & 0.5M & 86.9 & \textbf{98.5} & \underline{99.5} & \underline{74.6} & \underline{93.5} & \underline{96.6} & 62.7 & 86.1 & 92.1 & \underline{46.6} & \underline{73.8} & \underline{83.4} \\ 
AdaptFormer~\cite{chen2022adaptformer} & 0.5M & 87.4 & 97.6 & 99.1 & 73.9 & 92.6 & \textbf{96.7} & \underline{63.7} & \underline{86.2} & \underline{92.5} & 46.3 & \underline{73.8} & \underline{83.4} \\ 
\midrule
\textbf{Ours} & 0.5M & \textbf{89.6} & \underline{98.0} & \textbf{99.6} & \textbf{75.1} & \textbf{93.6} & \textbf{96.7} & \textbf{64.2} & \textbf{86.8} & \textbf{92.7} & \textbf{47.5} & \textbf{74.4} & \textbf{83.8} \\
\bottomrule
\end{tabular}
}
\label{tab:main_image}
\end{table}

Compared with adapter-based methods~\cite{he2021towards, chen2022adaptformer}, our method still surpasses them with fewer or comparable parameters.
For example, on Flickr30K dataset, our method excels beyond two adapter baselines by 2.7 and 2.2 on TR R@1. 
The main reason for this phenomenon is that baselines are designed for uni-modal tasks, \textit{e.g.}, image classification and English-to-Romanian translation. 
As a result, a crucial gap emerges, with no interaction occurring between vision and language modalities.

\begin{table}[!t]\scriptsize
\caption{Ablation study on the effectiveness of the weight-sharing mechanism across five datasets. \textit{w/o share} variant is implemented by inserting uni-modal adapters into all layers.}
\label{tab:ab_share}
\vspace{-10pt}
\begin{center}
\resizebox{0.8\columnwidth}{!}{%
\begin{tabular}{p{1.2cm} p{1.5cm}<{\centering} | p{1.8cm} p{1.8cm} p{1.8cm} | p{1.8cm}}
    \toprule
Dataset & Methods & $\text{R@1} \uparrow$ & $\text{R@5} \uparrow$ & $\text{R@10} \uparrow$ & $\text{Avg}$ \\
    \midrule
\multirow{2}{*}{MSR-VTT} & \text{w$/$o Share} & 44.2 & 70.6 & 79.6 & 64.8 \\
& \textbf{Ours} & \textbf{45.0} \textbf{\textcolor{mygreen}{(+0.8)}} & \textbf{72.3} \textbf{\textcolor{mygreen}{(+1.7)}} & \textbf{81.0} \textbf{\textcolor{mygreen}{(+1.4)}} & \textbf{66.1} \textbf{\textcolor{mygreen}{(+1.3)}} \\
    \midrule
\multirow{2}{*}{DiDeMo} & \text{w$/$o Share} & 43.8 & 70.8 & 80.6 & 65.1  \\
& \textbf{Ours} & \textbf{45.6} \textbf{\textcolor{mygreen}{(+1.6)}} & \textbf{71.8} \textbf{\textcolor{mygreen}{(+1.0)}} & \textbf{80.8} \textbf{\textcolor{mygreen}{(+0.2)}} & \textbf{66.1} \textbf{\textcolor{mygreen}{(+1.0)}} \\
    \midrule
\multirow{2}{*}{ActivityNet} & \text{w$/$o Share} & 39.8 & 70.2 & 82.4 & 64.1 \\
& \textbf{Ours} & \textbf{40.9} \textbf{\textcolor{mygreen}{(+1.1)}} & \textbf{71.6} \textbf{\textcolor{mygreen}{(+1.4)}} & \textbf{82.7} \textbf{\textcolor{mygreen}{(+0.3)}} & \textbf{65.1} \textbf{\textcolor{mygreen}{(+1.0)}}\\
    \midrule
\multirow{2}{*}{Flickr30K} & \text{w$/$o Share} & 87.4 & 97.6 & 99.5 & 94.8  \\
& \textbf{Ours} & \textbf{89.6} \textbf{\textcolor{mygreen}{(+2.2)}} & \textbf{98.0} \textbf{\textcolor{mygreen}{(+0.4)}} & \textbf{99.6} \textbf{\textcolor{mygreen}{(+0.1)}} & \textbf{95.7} \textbf{\textcolor{mygreen}{(+0.9)}}\\
    \midrule
\multirow{2}{*}{MSCOCO} & \text{w$/$o Share} & 63.6 & 86.3 & 92.1 & 80.6  \\
& \textbf{Ours} & \textbf{64.2} \textbf{\textcolor{mygreen}{(+0.6)}} & \textbf{86.8} \textbf{\textcolor{mygreen}{(+0.5)}} & \textbf{92.7} \textbf{\textcolor{mygreen}{(+0.6)}} & \textbf{81.2} \textbf{\textcolor{mygreen}{(+0.6)}} \\
    \bottomrule
\end{tabular}%
}
\end{center}
\vspace{-15pt}
\end{table}

\subsection{Ablation Study of Weight-sharing Mechanism}
\textbf{Effectiveness of weight-sharing across datasets.}
To demonstrate the effectiveness of our weight-sharing mechanism, we conduct thorough ablation studies on all five datasets. 
In \cref{tab:ab_share}, we show that the sharing weight not only boosts the video-text retrieval performance but also image-text retrieval. 
It indicates sharing weight facilitates the re-aligning of CLIP's vision and language feature space. 
Note that on MSR-VTT, our method achieves significant improvement on all metrics.
Furthermore, on DiDeMo and ActivityNet, our method surpasses the baseline significantly on the R@1/5 metrics and achieves better performance on R@10.
For image-text retrieval, our method boosts the R@1 performance on Flickr30K by 2.2 and consistently improve all metrics on MSCOCO.
In conclusion, sharing weight between vision and language encoders helps to unleash the power of the pre-trained CLIP model and boosts retrieval performance.
The take-away message is that encoder-level implicit cross-modal interaction is helpful for aligning features from different modalities.
We hope that this finding will inspire other researchers to investigate more efficient implicit cross-modal interactions in vision-language retrieval domain.

\begin{figure}[!tp]
  \centering
  \begin{minipage}[b]{0.28\textwidth}
    \includegraphics[width=\textwidth]{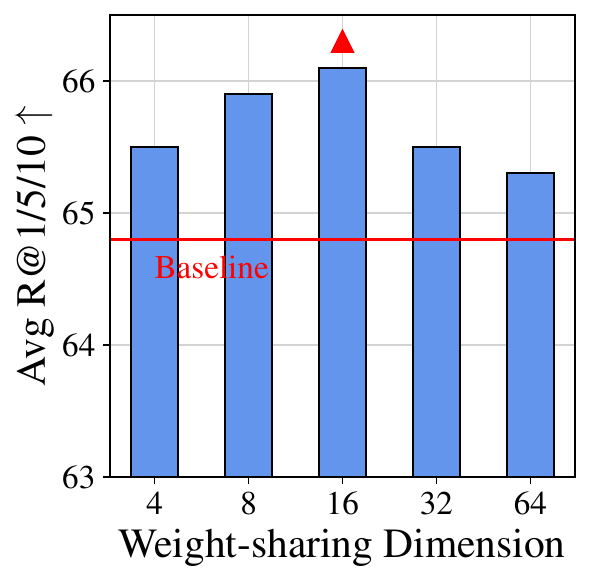}
    \caption{Ablation study.}
    \label{fig:ab_neckdim_sharedim}
  \end{minipage}
  \begin{minipage}[b]{0.6\textwidth}\scriptsize
        \centering
        \begin{tabular}{p{.6cm}<{\centering} p{.8cm}<{\centering} | p{.8cm}<{\centering} p{.6cm}<{\centering} p{.8cm}<{\centering} | p{1.7cm}}
                \toprule
             Down & Up & $\text{R@1} \uparrow$ & $\text{R@5} \uparrow$ & $\text{R@10} \uparrow$ & $\text{Avg}$ \\
                \midrule
            & & 44.2 & 70.6 & 79.6 & 64.8 \\
            \cmark & & 44.2 & 71.3 & 81.9 & 65.8 \\
            & \cmark & 45.0 & 72.3 & 81.0 & \textbf{66.1} \textbf{\textcolor{mygreen}{(+1.3)}} \\
            \cmark & \cmark & 44.5 & 71.3 & 81.8 & 65.9 \\
                \bottomrule
        \end{tabular}
        \captionof{table}{Ablation study on application positions inside adapter on MSR-VTT.}
        \label{tab:ab_sharepos}
    \end{minipage}
\end{figure}

\begin{table}[!t]\scriptsize
\caption{Verification of modality-specific feature learning of up and down-projection.}
\vspace{-10pt}
\begin{center}
\begin{tabular}{p{.6cm}<{\centering} | p{.6cm}<{\centering}  p{.6cm}<{\centering} | p{.6cm}<{\centering}  p{.6cm}<{\centering} | p{.6cm}<{\centering}  p{.6cm}<{\centering} | p{.6cm}<{\centering}  p{.6cm}<{\centering} | p{.6cm}<{\centering}  p{.6cm}<{\centering}}
    \toprule
 & \multicolumn{2}{c|}{Layer-2} & \multicolumn{2}{c|}{Layer-4} & \multicolumn{2}{c|}{Layer-6} & \multicolumn{2}{c|}{Layer-8} & \multicolumn{2}{c}{Layer-10} \\
& up & down & up & down & up & down & up & down & up & down \\
    \midrule
Acc & 74.96 & 61.13 & 57.96 & 53.46 & 87.79 & 68.67 & 97.96 & 85.38 & 98.79 & 98.54 \\
    \bottomrule
\end{tabular}
\end{center}
\vspace{-15pt}
\label{tab:ab_modality_specific}
\end{table}

\textbf{Effectiveness of weight-sharing across various dimensions.}
We further ablate the influence of changing the weight-sharing dimension $d_{s}$ on MSR-VTT. 
As shown in \cref{fig:ab_neckdim_sharedim}, the performance of the cross-modal adapter peaks at the $d_{s}$ of 16. 
Although performance begins to decline after exceeding 16, it remains superior to the baseline.
This indicates that our method is robust to the weight-sharing dimension, remaining effective across multiple hyper-parameter values.

\textbf{Effectiveness of weight-sharing across positions inside the cross-modal adapter.}
The weight-sharing mechanism can be deployed not only in up-projection layer but also down-projection layer.
To figure out which one is the best, we conduct an ablation study of different weight-sharing strategies in \cref{tab:ab_sharepos}.
We observe that weight-sharing is effective wherever (only up or down, or both) it is applied, i.e., w/ share variants surpass w/o share one on both datasets.
Furthermore, sharing weight in the up-projection layer outperforms all other strategies. 
The phenomenon may be due to the up-projection in the vanilla adapter learning more modality-specific features, which impair cross-modal alignment. 
Therefore, features in the up-projection may require weight-sharing to enhance cross-modality integration.
To validate our hypothesis, we collect output features from the vanilla adapters at various layers of both the vision and language encoders and train a classifier to categorize these features based on their originating modality.
The results in \cref{tab:ab_modality_specific} confirm our hypothesis that the up-projection layer tends to learn modality-specific features.

\begin{table}[!t]\scriptsize
\caption{Ablation study on various foundation models on Flickr30K.}
\vspace{-10pt}
\begin{center}
\begin{tabular}{p{1.5cm} p{1.5cm}<{\centering} | p{1.8cm} p{1cm} p{1cm}}
    \toprule
Model & Method & $\text{R@1} \uparrow$ & $\text{R@5} \uparrow$ & $\text{R@10} \uparrow$ \\
    \midrule
\multirow{2}{*}{ALBEF-4M} & w/o share & 93.3 & 99.0 & 99.9 \\
& \textbf{Ours} & \textbf{94.1} \textbf{\textcolor{mygreen}{(+0.8)}} & \textbf{99.2} & 99.9 \\
\multirow{2}{*}{ALBEF-14M} & w/o share & 94.8 & 99.5 & 99.9 \\
& \textbf{Ours} & \textbf{95.6} \textbf{\textcolor{mygreen}{(+0.8)}} & \textbf{99.6} & \textbf{100.0} \\
\multirow{2}{*}{BLIP-14M} & w/o share & 95.4 & 99.6 & 99.8 \\
& \textbf{Ours} & \textbf{96.0} \textbf{\textcolor{mygreen}{(+0.6)}} & 99.6 & \textbf{100.0} \\
\multirow{2}{*}{BLIP-129M} & w/o share & 96.5 & 99.9 & 100.0 \\
& \textbf{Ours} & \textbf{97.2} \textbf{\textcolor{mygreen}{(+0.6)}} & \textbf{100.0} & 100.0 \\
    \bottomrule
\end{tabular}
\end{center}
\label{tab:ab_diff_models}
\vspace{-15pt}
\end{table}

\textbf{Effectiveness of weight-sharing across foundation models.}
In this paper, most of our experiments were conducted using the pre-trained CLIP model. 
However, to clearly demonstrate that our approach isn't limited to a specific foundation model.
We further evaluate our method on ALBEF~\cite{li2021align} and BLIP~\cite{li2022blip} following their settings. 
Specifically, for BLIP, we insert the cross-modal adapters inside all layers of vision and language encoders. 
In the ALBEF architecture, the vision encoder has 12 layers while the language encoder has 6. 
Given the cross-modal adapter's need for inter-modality interaction, we integrated it into the first six layers of the vision encoder and all the layers of language encoder.
As we can see from \cref{tab:ab_diff_models}, our method consistently improves the R@1 across four models and achieves comparable or better performance on other metrics.
The findings highlight the effectiveness of the cross-modal adapter across different foundational models, underscoring the model-agnostic nature of our method.

\textbf{Extended training doesn't enhance w/o share variant's performance over weight-sharing.}
To validate the possibility that the suboptimal performance of the w/o share variant might stem from its prolonged convergence time, we decided to extend the training duration, testing it over longer epochs of 10 and 15.
As evidenced in \cref{tab:ab_extend_epoch}, even when trained for fewer epochs, the weight-sharing variant consistently surpasses the w/o share version, registering a lead of at least 0.7 on average R@K metric. 

\begin{table}[!t]\scriptsize
\caption{Ablation study of training duration on MSR-VTT.}
\vspace{-10pt}
\begin{center}
\begin{tabular}{p{1.cm} r | p{2cm} | p{1.cm}<{\centering} | p{1.cm} r | p{2cm} | p{1.cm}<{\centering}}
    \toprule
Methods & Epoch & $\text{R@1/5/10} \uparrow$ & $\text{Avg}$ & Methods & Epoch & $\text{R@1/5/10} \uparrow$ & $\text{Avg}$ \\
    \midrule 
\textbf{Ours} & 5 & \textbf{45.0}/\textbf{72.3}/81.0 & \textbf{66.1} & \multirow{2}{*}{w/o share} & 10 & 44.1/70.6/\textbf{81.4} & 65.4 \\
\text{w/o share} & 5 & 44.2/70.6/79.6 & 64.8                           & & 15 & 43.8/70.6/81.3 & 65.2 \\
    \bottomrule
\end{tabular}
\label{tab:ab_extend_epoch}
\end{center}
\vspace{-15pt}
\end{table}

\subsection{Ablation Study of Adapter Insertion Strategies}
\begin{figure}[!t]
    \centering
    \includegraphics[width=1.0\columnwidth]{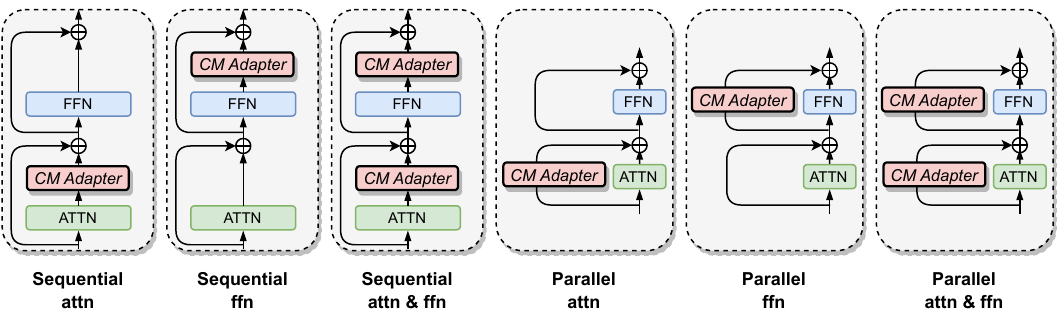}
    \caption{Diverse insertion forms and positions.}
    \label{fig: insertion_forms_and_positions}
    \vspace{-5pt}
\end{figure}

\textbf{Inserting adapters with different forms and positions inside encoders.} 
Previous works~\cite{houlsby2019parameter, he2021towards} shows the position and form of adapter insertion has a significant impact on performance. 
Houlsby et al.~\cite{houlsby2019parameter} places two adapters sequentially after both the self-attention (ATTN) and feedforward (FFN) sub-layer. 
Later, He et al.~\cite{he2021towards} designs a efficient adapter that is inserted only after ATTN sub-layer in a parallel way. 
Therefore, we perform an ablation study, as presented in \cref{tab:ab_diff_insert}, to evaluate the effectiveness of various insertion strategies (\cref{fig: insertion_forms_and_positions}).
First, we can observe that it is better to insert adapter in FFN rather than ATTN layer. 
Second, inserting adapters in both ATTN and FFN in a parallel way does not improve performance, while all metrics are greatly improved when adapters are inserted sequentially. 
As a result of these findings, we implement the cross-modal adapters in a sequential order across both the ATTN and FFN layers.

\begin{table}[!t]\scriptsize
\caption{Various insertion forms and positions. The row in gray represents final design.}
\vspace{-10pt}
\label{tab:ab_diff_insert}
\begin{center}
\begin{tabular}{p{1.5cm} p{1.7cm}<{\centering} | p{.8cm}<{\centering} p{.8cm}<{\centering} p{.8cm}<{\centering} | p{1.cm}<{\centering}}
\toprule
Form & Position & $\text{R@1} \uparrow$ & $\text{R@5} \uparrow$ & $\text{R@10} \uparrow$ & $\text{Avg}$ \\
\midrule
\multirow{3}{*}{Sequential} & \text{attn} & 42.0 & 68.3 & 78.8 & 63.0 \\
 & \text{ffn} & 43.2 & 69.0 & 80.1 & 64.1 \\
 & \cellcolor{Gray}\text{attn \& ffn} & \cellcolor{Gray}\textbf{45.0} & \cellcolor{Gray}\textbf{72.3} & \cellcolor{Gray}\textbf{81.0} & \cellcolor{Gray}\textbf{66.1} \\
\midrule
 \multirow{3}{*}{Parallel} & \text{attn} & 43.3 & 69.9 & 79.0 & 64.1  \\
 & \text{ffn} & 44.0 & 70.2 & 80.1 & 64.8 \\
 & \text{attn $\&$ ffn} & 43.8 & 70.3 & 80.8 & 65.0 \\
\bottomrule
\end{tabular}%
\end{center}
\vspace{-15pt}
\end{table}

\textbf{Inserting adapters into different modalities.}
In \cref{tab:ab_diff_modal}, we want to find out how inserting adapters into different modalities affects performance.
It can be seen that inserting adapters only in one modality will significantly reduce the performance. 
Specifically, their average R@K performance decrease by 1.4 and 2.4 when compared to the insertion of adapters in both modalities, as shown in the third row.
This observation suggests that for multi-modal tasks, it is imperative to implement adapters in both the vision and language encoders.
Furthermore, by fostering cross-modal interactions between the vision and language encoders, our method witnesses a boost in performance. 
Specifically, there's a notable increase of 1.3, as evidenced in the final two rows.

\begin{table}[!t]\scriptsize
\caption{Inserting cross-modal adapters into different modalities on MSR-VTT. $\mathit{V}$, $\mathit{L}$, and $\mathit{C}$ denote visual modality, linguistic modality, and cross-modal interaction.}
\vspace{-10pt}
\label{tab:ab_diff_modal}
\begin{center}
\begin{tabular}{p{.4cm}<{\centering}  p{.4cm}<{\centering} p{.4cm}<{\centering} | p{.8cm}<{\centering} p{.8cm}<{\centering} p{.8cm}<{\centering} | p{1.7cm}}
    \toprule
$\mathit{V}$ & $\mathit{L}$ & $\mathit{C}$ & $\text{R@1} \uparrow$ & $\text{R@5} \uparrow$ & $\text{R@10} \uparrow$ & $\text{Avg}$ \\
    \midrule
\cmark & & & 42.6 & 67.1 & 77.6 & 62.4 \\
& \cmark & & 41.1 & 69.8 & 79.4 & 63.4 \textbf{\textcolor{mygreen}{(+1.0)}} \\
\cmark & \cmark & & 44.2 & 70.6 & 79.6 & 64.8 \textbf{\textcolor{mygreen}{(+2.4)}} \\
\cmark & \cmark & \cmark &  \textbf{45.0} & \textbf{72.3} & \textbf{81.0} & \textbf{66.1} \textbf{\textcolor{mygreen}{(+3.7)}} \\
    \bottomrule
\end{tabular}
\end{center}
\vspace{-15pt}
\end{table}

\begin{table}[!t]\scriptsize
\caption{Inserting cross-modal adapters into different layers on MSR-VTT.} 
\vspace{-10pt}
\begin{center}
\begin{tabular}{p{4.5cm} | p{.8cm}<{\centering} p{.8cm}<{\centering} p{.8cm}<{\centering} | p{1.8cm}}
    \toprule
Methods (first / last N layers) & $\text{R@1} \uparrow$ & $\text{R@5} \uparrow$ & $\text{R@10} \uparrow$ & $\text{Avg}$ \\
    \midrule
12 Uni-Adapter (w/o share)   & 44.2 & 70.6 & 79.6 & 64.8 \\
\text{ }\textcolor{mygreen}{3 CM-Adapter} / 9 Uni-Adapter & 45.2 & 71.5 & 80.8 & \textbf{65.8} \textbf{\textcolor{mygreen}{(+1.0)}}\\
\text{ }\textcolor{mygreen}{6 CM-Adapter} / 6 Uni-Adapter & 45.1 & 69.8 & 81.5 & \textbf{65.5} \textbf{\textcolor{mygreen}{(+0.7)}}\\
\text{ }6 Uni-Adapter / \textcolor{mygreen}{6 CM-Adapter} & 45.4 & 71.9 & 82.3 & \textbf{66.5} \textbf{\textcolor{mygreen}{(+1.7)}}\\
\text{ }9 Uni-Adapter / \textcolor{mygreen}{3 CM-Adapter} & 44.5 & 71.9 & 81.4 & \textbf{65.9} \textbf{\textcolor{mygreen}{(+1.1)}}\\
\textcolor{mygreen}{12 CM-Adapter} & 45.0 & 72.3 & 81.0 & \textbf{66.1} \textbf{\textcolor{mygreen}{(+1.3)}}\\
    \bottomrule
\end{tabular}
\label{tab:ab_across_layers}
\end{center}
\vspace{-15pt}
\end{table}

\textbf{Inserting adapters into different layers.} 
Beginning with a variant devoid of weight-sharing components (first row in \cref{tab:ab_across_layers}), we progressively substituted the uni-modal adapter with our proposed cross-modal adapter, moving from the initial layers to the final ones.
As we can see from \cref{tab:ab_across_layers}, irrespective of the quantity or position of layers replaced, the cross-modal adapter consistently enhances the model's performance.
Especially, replacing the last six layers with cross-modal adapters achieves the best performance which significantly surpasses the baseline (first row) by 1.7 on average R@K.
Note that all variants utilizing cross-modal adapters represent a specific implementation of our method.
For the simplicity of the method, we insert cross-modal adapters to all layers of encoders by default.

\begin{figure}[!t]
\centering
\includegraphics[scale=0.613]{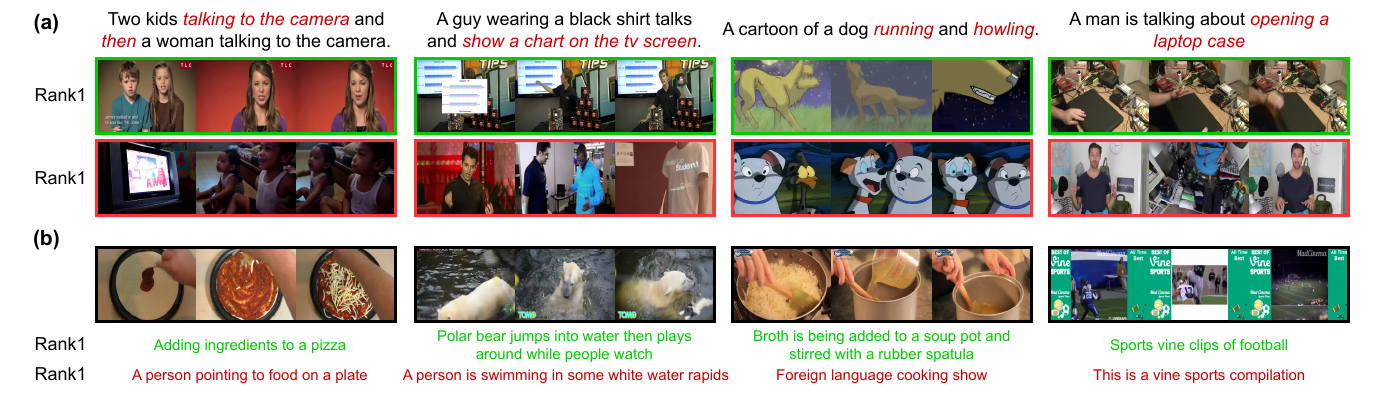}
\caption{
Visualizations results. 
\textcolor{mygreen}{Green} boxes show the rank-1 correct retrieval results of our method. 
\textcolor{red}{Red} boxes are baseline's wrong retrieval results.
}
\label{fig:visualization}
\centering
\end{figure}

\begin{figure}[!t]
\centering
\includegraphics[scale=0.64]{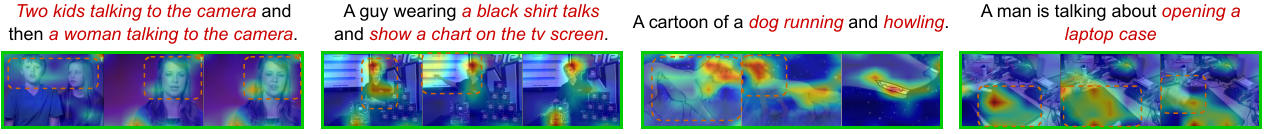}
\caption{
Visualizations of model's attention on frames. We highlight discriminative feature in both text and images.
}
\label{fig:visualization_attn}
\centering
\vspace{-15pt}
\end{figure}

\subsection{Visualization Results}
We visualize multiple text-to-video and video-to-text retrieval results on the MSR-VTT test set in \cref{fig:visualization}. 
From our visualizations, it's evident that our approach demonstrates a superior understanding of temporal dynamics, nuanced actions, and the intricate details of local visual elements, indicating better cross-modal alignment.
First, focusing on the example in the \cref{fig:visualization} (a) leftmost , our method adeptly discerns the temporal dynamics and actions.
Additionally, focusing on the third case in \cref{fig:visualization} (b), our method correctly understands the actions and objects in the video.
Furthermore, we visualize our model's attention weight on each frame in \cref{fig:visualization_attn}.
Clearly, our model successfully focuses on the key distinguishing features in the image.

\begin{figure}[!tp]
  \centering
  \begin{minipage}[b]{0.45\textwidth}
    \includegraphics[width=\textwidth]{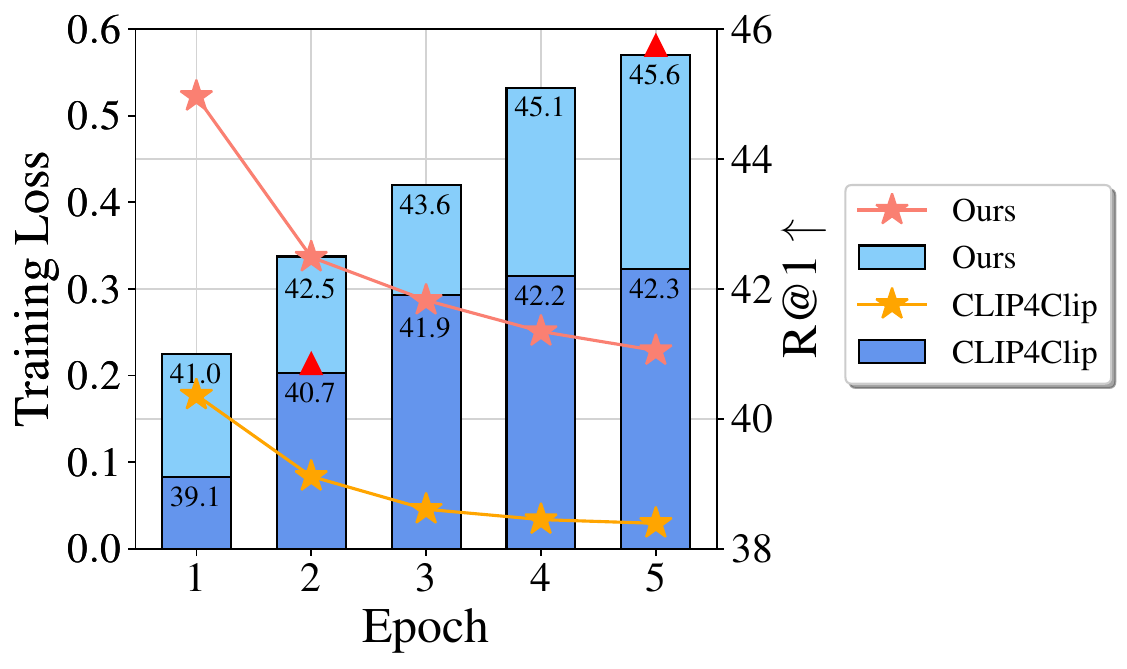}
    \vspace{-1.2cm}
    \caption{Overfitting phenomenon.}
    \label{fig:ab_overfit}
  \end{minipage}
  \begin{minipage}[b]{0.5\textwidth}\scriptsize
        \centering
        \begin{tabular}{l l l}
            \toprule
            Pretrain on & Method & Avg \\
            \midrule
            \multirow{2}{*}{DiDeMo} & Full-tuning~\cite{luo2021clip4clip} & 54.2\\
            & \textbf{Ours} & \textbf{55.3} \textbf{\textcolor{mygreen}{(+1.1)}} \\
            \midrule
            \multirow{2}{*}{ActivityNet} & Full-tuning~\cite{luo2021clip4clip} & 53.9 \\
            & \textbf{Ours} & \textbf{55.6} \textbf{\textcolor{mygreen}{(+1.7)}} \\
            \bottomrule
        \end{tabular}
        \captionof{table}{Generalization test on MSR-VTT.}
        \label{tab:ab_overfit}
    \end{minipage}
\vspace{-15pt}
\end{figure}

\subsection{Overfitting Phenomenon of Full-tuning method.}
We investigate the training process of full-tuning method~\cite{luo2021clip4clip} and our approach on DiDeMo which has smallest training set.
As we can see from \cref{fig:ab_overfit}, the training loss of the full-tuning method decreases rapidly due to its large capacity while our method's loss declines much slower.
However, lower training loss does not equate to better performance, indicating a severe overfitting phenomenon.
Since large models are data-hungry, DiDeMo with only 8k training samples may not be adequate for fine-tuning a model with more than 123M parameters.
On the contrary, our method minimizes trainable parameters to under 1\% of the original, thus mitigating overfitting.
We further validated the models pre-trained on DiDeMo and ActivityNet using MSR-VTT to test their generalizability in \cref{tab:ab_overfit}.
Clearly, our method learns more generalizable features.
Moreover, in \cref{tab:main_actnet}, our method matches the full-tuning approach in performance but shows better generalization on MSR-VTT (\cref{tab:ab_overfit}), hinting at overfitting in the full-tuning method.

\subsection{Efficiency of Cross-Modal Adapter}
\begin{figure}[!t]
    \centering
    \includegraphics[width=0.7\columnwidth]{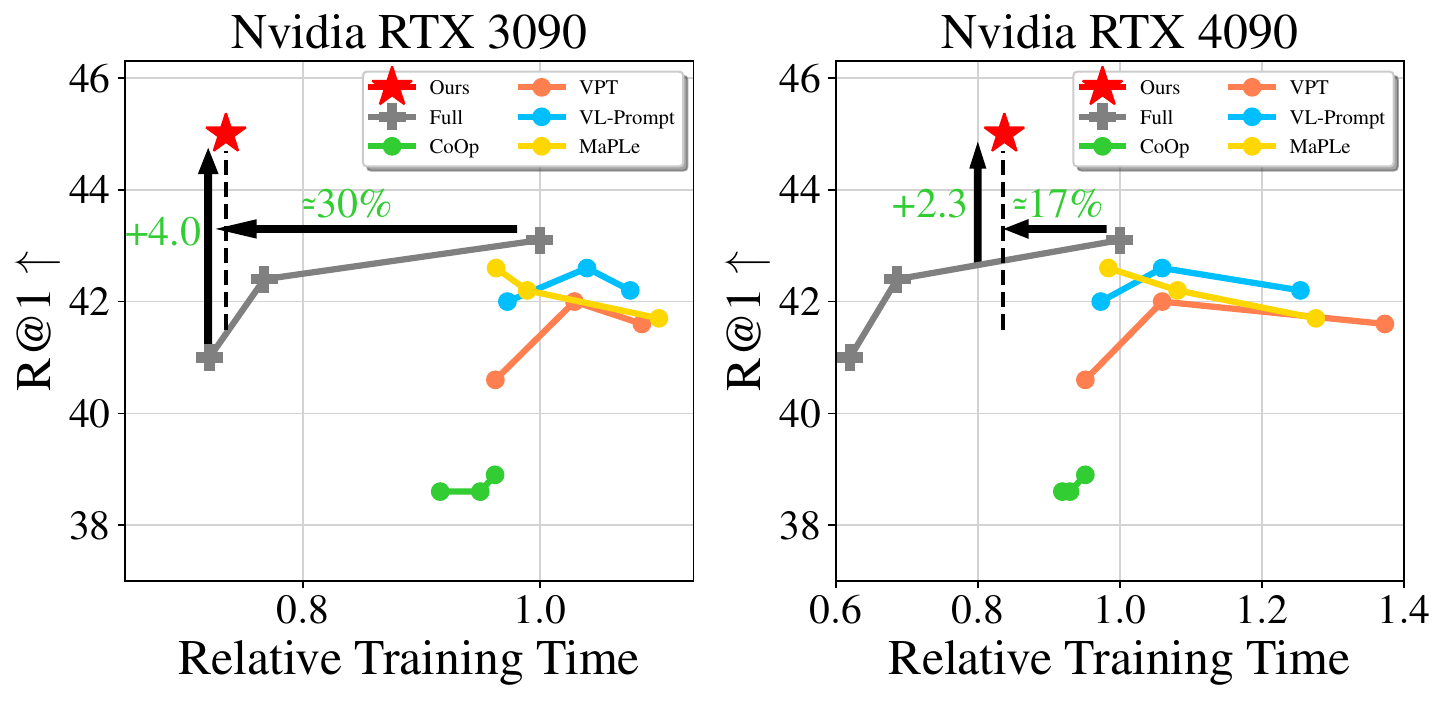}
    \vspace{-0.3cm}
    \caption{Efficiency of the cross-modal adapter.}
    \label{fig:efficiency}
    \vspace{-15pt}
\end{figure}

To showcase the efficiency of proposed method, we conduct empirical tests on its training speeds and memory consumption using 8 RTX-3090 and RTX-4090 GPUs.
As illustrated in \cref{fig:efficiency}, our approach presents a more favorable balance between training speed and overall performance.
Specifically, compared to the full-tuning method, our approach boosts performance by 4.0 and 2.3 under the same training time.
Furthermore, our model surpasses the best full-tuning model while also reducing training time by 30\% and 17\%, proving its higher efficiency.

%% file: 2nd_review_revise/5_conclusion.tex
\section{Conclusion}
In this paper, we explored adapter-based parameter-efficient transfer learning for the vision-language retrieval tasks. 
To address the challenge of introducing feature-level interaction that causes a surge in computational cost, we proposed a novel \namelong{} to enable encoder-level implicit cross-modal interaction via a weight-sharing mechanism. 
Experiments showed that our approach is robust to different foundation models and robust to the hyper-parameters.
Furthermore, our approach surpassed prompt-based methods significantly, indicating that the adapters may offer a better parameter-efficient solution for transferring pre-trained foundation models to vision-language retrieval tasks.

Although experiments have validated the effectiveness of our idea, it should be noted that there still exist several limitations. 
Firstly, the weight-sharing mechanism is specifically designed for the adapter, and applying it to other parameter-efficient transfer learning methods, such as LoRA~\cite{hu2021lora}, may pose a challenge. 
Secondly, our work concentrates on promoting alignment at the encoder level. Consequently, we did not explore a parameter-efficient vision-language matching module that occurs post-encoder, which might further facilitate cross-modal alignment. 
We believe these limitations are also promising research directions that are worth further investigation.